\documentclass[10pt,twocolumn,letterpaper]{article}

\usepackage[pagenumbers]{iccv} 

\definecolor{iccvblue}{rgb}{0.21,0.49,0.74}
\usepackage{multirow}
\usepackage{pifont}
\usepackage{makecell}
\usepackage{listings}
\usepackage[pagebackref,breaklinks,colorlinks,allcolors=iccvblue]{hyperref}

\def\confName{ICCV}
\def\confYear{2025}

\def\ours{PlanGen}
\def\hico{HiCo}

\def\qwenvl{Qwen-VL-Chat}
\def\cogvlmg{CogVLM-grounding}
\def\gdino{Grounding-DINO}

\newcommand{\gray}[1]{\textcolor{gray}{#1}}

\title{PlanGen: Towards Unified Layout Planning and Image Generation \\ in Auto-Regressive Vision Language Models}

\author{Runze He \quad
Bo cheng \quad
Yuhang Ma \quad
Qingxiang Jia \quad
Shanyuan Liu \quad\\
Ao Ma\quad
Xiaoyu Wu\quad
Liebucha Wu\quad
Dawei Leng \footnotemark[2] \quad
Yuhui Yin\quad \\
\texttt{360 AI Research} \quad \\
\small
\texttt{herunze8265@gmail.com,} \texttt{\{chengbo1, lengdawei\}@360.cn}\\
}

\begin{document}

\twocolumn[{
\renewcommand\twocolumn[1][]{#1}
\maketitle
\begin{center}
    \centering
    \vspace*{-.6cm}
    \includegraphics[width=1.0\textwidth]{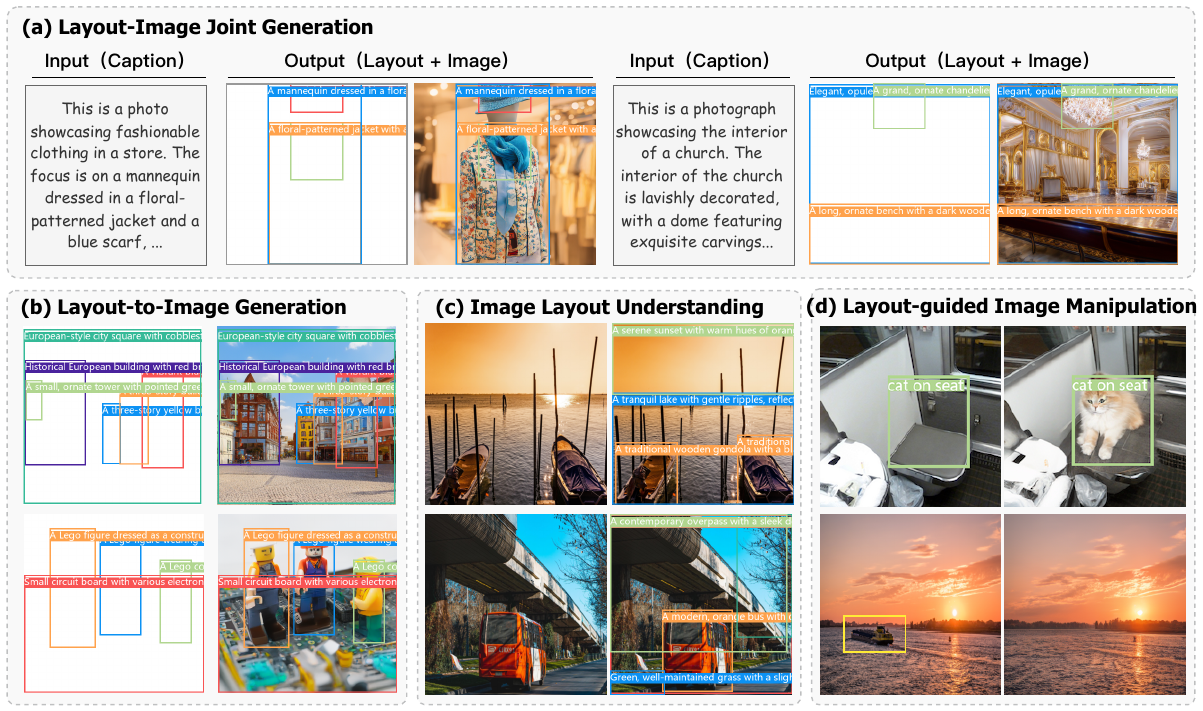}
    \vspace*{-.2cm}
    \captionof{figure}{\ours{} models \textbf{layout planning} and \textbf{image generation} jointly, allowing layout planning before generating corresponding images, and the two processes are completed in a unified model.
    \ours{} can perform multi-type tasks related to layout, including \textbf{a)} layout-image joint generation, \textbf{b)} layout-to-image generation, \textbf{c)} image layout understanding and \textbf{d)} layout-guided image manipulation.}
\label{fig:fig1}
\end{center}
}]

\footnotetext[2]{Corresponding authors.}

\begin{abstract}
In this paper, we propose a unified layout planning and image generation model, \ours{}, which can pre-plan spatial layout conditions before generating images.
Unlike previous diffusion-based models that treat layout planning and layout-to-image as two separate models, \ours{} jointly models the two tasks into one autoregressive transformer using only next-token prediction.
\ours{} integrates layout conditions into the model as context without requiring specialized encoding of local captions and bounding box coordinates, which provides significant advantages over the previous embed-and-pool operations on layout conditions, particularly when dealing with complex layouts.
Unified prompting allows \ours{} to perform multitasking training related to layout, including layout planning, layout-to-image generation, image layout understanding, etc.
In addition, \ours{} can be seamlessly expanded to layout-guided image manipulation thanks to the well-designed modeling, with teacher-forcing content manipulation policy and negative layout guidance.
Extensive experiments verify the effectiveness of our \ours{} in multiple layout-related tasks, showing its great potential.
Code is available at: \url{https://360cvgroup.github.io/PlanGen}.

\end{abstract}

\section{Introduction}
In recent years, the field of image synthesis has received widespread attention from the community, and technologies such as GAN~\cite{goodfellow2020generative,karras2019style,zhang2022styleswin}, autoregressive image generation~\cite{sun2024autoregressive_llamagen, VAR, Infinity, li2025autoregressive_mar}, and diffusion models~\cite{song2021ddim,ho2020ddpm} have made great progress. Among them, diffusion models dominate because of their more stable training, sampling diversity, and high quality. Recently, autoregressive image generation has been widely explored because of its scalability and compatibility with large language models (LLMs).

\begin{figure}[!t]
\centering
\includegraphics[width=3.2in]{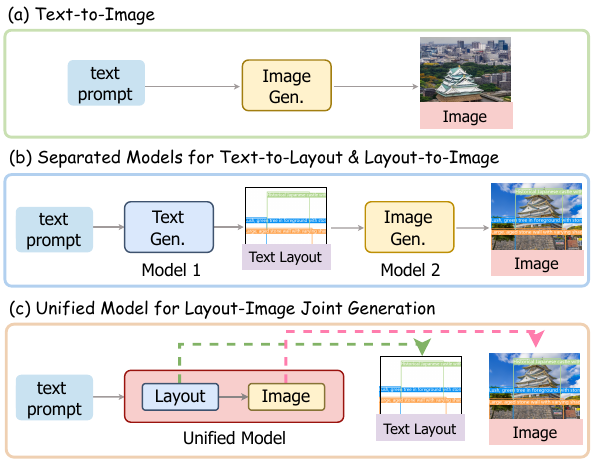}
\caption{Different paradigms of image generation. \textbf{a)} The naive text-to-image cannot control the layout of the generated images. \textbf{b)} Previous methods use two independent models to complete layout planning and image generation. \textbf{c)} We adopt a unified model to complete layout planning and image generation.}
\label{fig:task}
\vspace{-15pt}
\end{figure}

With the continuous advancement of text-to-image foundation models~\cite{ramesh2022dalle2,betker2023dalle3,saharia2022imagen,GLIDE} , their applications in commercial scenarios have become increasingly diverse. While fundamental text-to-image models primarily focus on text-image alignment, users often have more specific requirements regarding the spatial arrangement and interactions between objects.
As shown in Figure~\ref{fig:task}, recent diffusion-based methods~\cite{li2023gligen,wang2024instancediffusion,zhou2024migc,cheng2024hico,zhang2024creatilayout} introduce layout conditions into the image generation process. Despite the promising results, it faces two shortcomings: 1) Inability to expand to layout-image joint generation, which leads to reliance on another layout planning model. 2) Suboptimal encoding of layout condition, which is mainly caused by embed-and-pool operation on local captions to unify layout conditions for facilitating model training.

We propose an autoregressive visual-language model, \ours{}, which can complete layout planning and layout-to-image generation in a unified model. Just like thinking about what object each area should be before generating an image, such an explicit planning process allows the model to enjoy more powerful image generation capabilities.
\ours{} takes layout conditions as context input, with extremely strong flexibility and scalability, no need to compress the layout conditions as previous methods~\cite{li2023gligen, wang2024instancediffusion, zhou2024migc, zhang2024creatilayout}, allowing as detailed and complex local descriptions as possible, leading to a more aligned layout-to-image generation thanks to transformer's long context dependencies.
\ours{} allows multi-task training related to layout through unified prompting, including layout planning, layout-to-image generation, and image layout understanding. Additionally, through teacher-forcing content manipulation and negative layout guidance, \ours{} can be extended to layout-guided image manipulation without specialized training.

Our contributions are summarized as follows: (1) We present the first layout-image joint generation model based on the autoregressive vision-language model. (2) We propose a unified prompting paradigm to allow layout-image joint modeling, as well as multitasking training related to layout conditions. (3) We further extend \ours{} to layout-guided image manipulation through teacher-forcing content manipulation and negative layout guidance.

\section{Related Works}
\noindent \textbf{Vision-Language Models (VLMs)}
In recent years, large vision-language models~\cite{llama3.1,hui2024qwen2.5-paper,alayrac2022flamingo,achiam2023gpt4-turbo, Brown2020GPT3} have made remarkable breakthroughs.
As a critical research direction, unified model with understanding and vision generation abilities~\cite{xie2024showo,zhou2024transfusion,zhao2024monoformer,wang2024emu3, ge2024seedx,wu2024janus,chen2025janus_pro,team2024chameleon,fang2024puma} is being widely explored.
Early approach SEED-X~\cite{ge2024seedx} applies another independent diffusion model to complete the visual generation task.
Show-o~\cite{xie2024showo} performs autoregressive and discrete diffusion modeling in one transformer.
Transfusion~\cite{zhou2024transfusion} and MonoFormer~\cite{zhao2024monoformer} integrate the diffusion process into next-token prediction.
Emu3~\cite{wang2024emu3}, Janus~\cite{wu2024janus} as well as Janus-Pro~\cite{chen2025janus_pro} only use next-token prediction in training and also achieve promising results.



\noindent \textbf{Autoregressive Image Generation}
Autoregressive image generation models~\cite{sun2024autoregressive_llamagen, VAR, Infinity, tang2024hart, li2025autoregressive_mar, sun2024xpromptuniversalincontextimage} have recently received widespread attention from the community due to their scalability and compatibility with LLMs.
LLamaGen~\cite{sun2024autoregressive_llamagen} apply the original ``next-token prediction'' paradigm of LLMs to the visual generation domain.
MaskGIT~\cite{chang2022maskgit} introduces a bidirectional transformer decoder and predicts randomly masked tokens by attending to tokens in all directions, instead of decoding images sequentially following the raster scan ordering.
VAR~\cite{VAR} redefines the autoregressive image generation as a coarse-to-fine process, i.e.``next-scale prediction'', showing promising results.
MAR~\cite{li2025autoregressive_mar} performs autoregressive image generation in a continuous-valued space, with additional diffusion MLPs to model the per-token probability distribution.
 

\noindent \textbf{Layout-to-Image Generation}
The layout-to-image generation methods can generally be categorized into training-free~\cite{chen2024training_cag, xie2023boxdiff} and training-based~\cite{feng2024ranni, cheng2024hico, zhang2024creatilayout, li2023gligen, wang2024instancediffusion, zhou2024migc, gu2024kaleido, nie2024compositional_blobgen} approaches. Early methods, such as CAG~\cite{chen2024training_cag}, which do not require training, laid the foundation for the field. Recent research, however, has primarily focused on training-based approaches. For instance, GLIGEN~\cite{li2023gligen} explored specialized attention mechanisms, while HiCo~\cite{cheng2024hico} investigated methods for controlling the introduction of conditions, which can be used seamlessly as plugins. Furthermore, the introduction of CreatiLayout~\cite{zhang2024creatilayout} in MM-DiT~\cite{esser2024sd3, flux} has further advanced controllable image generation based on layout design.

\noindent \textbf{Image Manipulation}
The huge advances in image generation have also promoted research on image manipulation, i.e. image editing. The task is also dominated by diffusion-based models.
Early training-free methods~\cite{Hertz2022Prompt2prompt, mokady2023null, cao_2023_masactrl, Tumanyan_2023_CVPR_pnp}, such as Prompt-to-Prompt~\cite{Hertz2022Prompt2prompt}, complete the image editing process by manipulating attention mechanisms. To edit real images, it is also necessary to inversion the real images, and the editing effect is not ideal.
Some methods~\cite{avrahami2024diffuhaul, mou2023dragondiffusion, mou2023diffeditor, epstein2023diffusion_selfgudance} have been proposed to achieve diversified editing like object dragging.
Instruction-based methods~\cite{brooks2023instructpix2pix, Zhang2023MagicBrush, fu2024mgie, huang2023smartedit, zhao2024ultraedit, he2024freeedit} represented by Instruct-Pix2Pix~\cite{brooks2023instructpix2pix} use natural language instructions to complete image editing, which relies on training on large-scale instruction editing datasets.



\section{Methods}
\begin{figure*}[!t]
\centering
\includegraphics[width=7in]{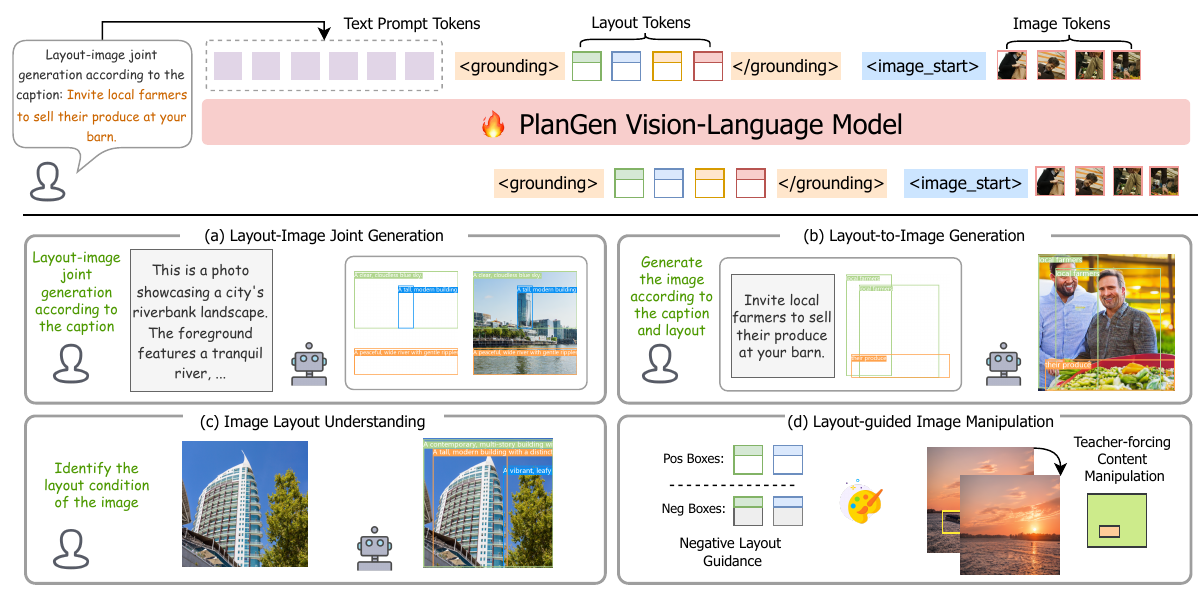}
\caption{\textbf{Upper:} \ours{} models layout planning and image generation jointly in an autoregressive visual-language model through a unified prompting design with the next-token prediction training objective. \textbf{Lower:} Illustration of \ours{}’s multitasking related to layout: \textbf{a)} layout-image joint generation, \textbf{b)} layout to image generation, \textbf{c)} image layout understanding and \textbf{d)} layout-guided image manipulation.}
\label{fig:pipeline}
\vspace{-10pt}
\end{figure*}

\subsection{Overview}
As shown in Figure~\ref{fig:pipeline}, \ours{} based on the architecture of the autoregressive vision language model, supports layout-image joint generation in one transformer with causal attention.
Given a text prompt $T$ by the user, instead of directly generating the image, \ours{} performs spatial layout planning first to obtain the layout condition $C$, which imagines the desired position of each subject in the text prompt, and then completes the image generation process under the condition of both text prompt $T$ and planned layout $C$.

\subsection{Joint Generation Modeling}

\subsubsection{Unified Prompting}
\label{sec:uni_prompt}
\begin{figure}[!t]
\centering
\includegraphics[width=3.2in]{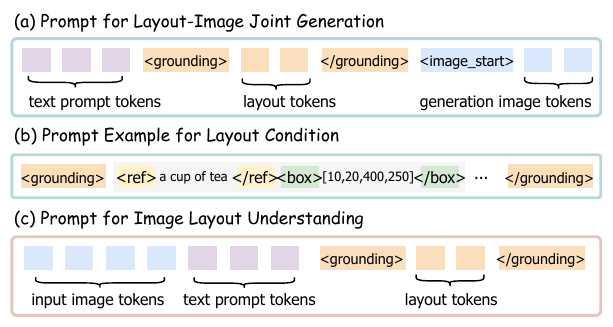}
\caption{Prompt design of \ours{}. \textbf{a)} Prompt for Layout-Image Joint Generation. \textbf{b)} Prompt Example for Layout Condition. \textbf{c)} Prompt for Image Layout Understanding.}
\label{fig:prompt}
\vspace{-10pt}
\end{figure}


As shown in Figure~\ref{fig:prompt}a, the layout planning and image synthesis of \ours{} are performed in sequence during the same round of inference. We use newly inserted special tokens \texttt{<grounding>} and \texttt{</grounding>} to mark the beginning and the end of the layout condition. After the layout condition is planned, we use \texttt {<image\_start>} to trigger the image generation process.

To help the model understand the layout condition, as shown in Figure~\ref{fig:prompt}b, we use \texttt{<ref>} and \texttt{</ref>} to wrap local descriptions like Qwen~\cite{Qwen-VL}. Each local description corresponds to a bounding box coordinate list, which is placed between \texttt{<box>} and \texttt{</box>}. The coordinate list contains four integers standardized to 0-1000, representing the upper left corner and the lower right corner of the bounding box.

Thanks to the flexible context length in autoregressive models, local descriptions in \ours{} can be as detailed as needed, in contrast to the previous diffusion-based methods~\cite{li2023gligen,zhang2024creatilayout,zhou2024migc,wang2024instancediffusion} which need to perform pooling or truncation operations to reduce and unify the length of the local description embeddings. In addition, due to the parallel training advantages of autoregressive transformers, layout planning and layout-to-image generation can perform training in the same batch.



\subsubsection{Image Layout Understanding Learning}
\label{sec:mmu}
The ability to understand the layout of real images can help the model generate images that are more in line with the layout conditions, intuitively, because this can further strengthen the model's in-depth understanding of the relationship between layout conditions and corresponding images. In addition, including the task of image layout understanding to \ours{} can move further towards a more general layout VLM.

Given an image $I$, we expect \ours{} to be able to analyze its layout conditions. 
Specifically, we first extract high-dimensional image features from $I$ using SigLIP~\cite{zhai2023SigLIP} encoder, then flatten and map them into the text feature space, following Janus-Pro~\cite{chen2025janus_pro}.
The prompt paradigm of image layout understanding is illustrated in Figure~\ref{fig:prompt}c.
In order to reduce the difficulty of directly predicting the layout conditions, we predict the caption first before predicting the layout conditions.

\subsubsection{Training Objectives}

The training of all tasks in \ours{} is based on next-token prediction.
We assume a sequence with $M$ text prompt tokens $\mathbf{t}=\{\mathbf{t_1},\mathbf{t_2},...,\mathbf{t_M}\}$, $N$ layout tokens $\mathbf{l}=\{\mathbf{l_1},\mathbf{l_2},...,\mathbf{l_N}\}$, $X$ image tokens to be generated $\mathbf{g}=\{\mathbf{g_1},\mathbf{g_2},...,\mathbf{g_X}\}$.
Without loss of generality, we introduce layout planning and layout-to-image generation that can be trained in parallel separately.

\noindent\textbf{Layout Planning}
We maximize the likelihood of layout tokens given text prompt tokens and all previously generated layout tokens by employing the standard language modeling objective:
\begin{equation}
    \mathcal{L}_{LP} = \sum_{i} \text{log} p_\theta(\mathbf{l_i} | \mathbf{t}, \mathbf{l_{<i}}),
\end{equation} 
where $p$ indicates the conditional probability of \ours{}, parameterized by weights $\theta$.

\noindent\textbf{Layout-to-Image Generation}
Similarly, we maximize the likelihood of image tokens given text prompt tokens, layout tokens, and previously generated image tokens, as formulated below:
\begin{equation}
    \mathcal{L}_{LIG} = \sum_{i} \text{log} p_\theta(\mathbf{g_i} | \mathbf{t}, \mathbf{l}, \mathbf{g_{<i}})  
\end{equation} 

\noindent\textbf{Image Layout Understanding}
The task requires the introduction of $K$ conditional image tokens $\mathbf{q}=\{\mathbf{q_1},\mathbf{q_2},...,\mathbf{q_K}\}$.
We maximize the likelihood of layout tokens as follows:
\begin{equation}
    \mathcal{L}_{ILU} = \sum_{i} \text{log} p_\theta(\mathbf{l_i} | \mathbf{q}, \mathbf{t}, \mathbf{l_{<i}})
\end{equation} 

The total training loss is summarized as follows:

\begin{equation}
    \mathcal{L} = \alpha \mathcal{L}_\text{LP} + 
    \beta \mathcal{L}_\text{LIG} + \gamma \mathcal{L}_\text{ILU},
\end{equation}
where $\alpha$, $\beta$, $\gamma$ are the hyperparameters to balance different losses, which are all set to 1 empirically in our experiments.

\subsection{Layout-guided Image Manipulation}
Benefiting from the modeling paradigm defined by layout conditions, \ours{} could control the generation of the contents of the local area based on the corresponding local captions and bounding boxes, which allows us to easily extend \ours{} to layout-guided image manipulation without further task-specific training.


\subsubsection{Teacher-forcing Content Manipulate}
One difficulty in image manipulation is to keep non-edited areas unchanged, and previous diffusion model-based approaches often require targeted training or inverse processes to achieve this goal, as its implicit features are difficult to manipulate directly.
In the paradigm of autoregressive image modeling, we can achieve this through teacher-forcing. 

Specifically, we can pre-calculate token position set $\mathcal{P}_{edit}$ based on the bounding boxes to be edited. For tokens within $\mathcal{P}_{edit}$ that require editing, we perform layout-to-image token sampling $\mathcal{F}$, or else we directly select pre-calculated tokens at corresponding mapped positions. The prediction process of the i-th image token $e_i$ is defined as follows:

\begin{equation}
    \mathbf{e_i} = \begin{cases} 
   \mathcal{F}(\mathbf{t},\mathbf{l},\mathbf{e_{<i}},i) & \text{if } i \in \mathcal{P}_\text{edit}, \\
   \mathcal{M}(i, \mathbf{o}) & \text{if } i \notin \mathcal{P}_\text{edit},
  \end{cases}
  \label{eq:tf}
\end{equation}
where $\mathbf{o}$ represents the image tokens of the original image, and $\mathcal{M}$ maps the i-token to corresponding image token in $o$.

\subsubsection{Negative Layout Guidance}
To further improve editing quality, we propose negative layout guidance to unlock the huge potential of layout-based image manipulation methods.
Existing methods often struggle to solve the artifacts after object deletion, as previous methods~\cite{winter2024objectdrop, sun2024attentiveeraser} have pointed out.

Thanks to our token-based layout conditions, we can easily incorporate negative layout guidance into the classifier-free guidance to suppress the generation of artifacts without causing additional inference overhead.

\begin{equation}
    \tilde{e_{\theta}}(\mathbf{t}, \mathbf{l}, \mathbf{g_{<i}}) = e_{\theta}(\tilde{\mathbf{t}}, \tilde{\mathbf{l}}, \mathbf{g_{<i}}) + s \cdot (e_{\theta}(\mathbf{t}, \mathbf{l}, \mathbf{g_{<i}}) - e_{\theta}(\tilde{\mathbf{t}}, \tilde{\mathbf{l}}, \mathbf{g_{<i}})),
    \label{eq:cfg}
\end{equation}
where $e_{\theta}$ and $\tilde{e_{\theta}}$ indicates predicted logits and modulated final logits, $\tilde{\mathbf{t}}$ and $\tilde{\mathbf{l}}$ are negative text prompt tokens and negative layout tokens respectively. The guidance scale $s$ is set to 5.

\begin{figure*}[!t]
\centering
\includegraphics[width=7in]{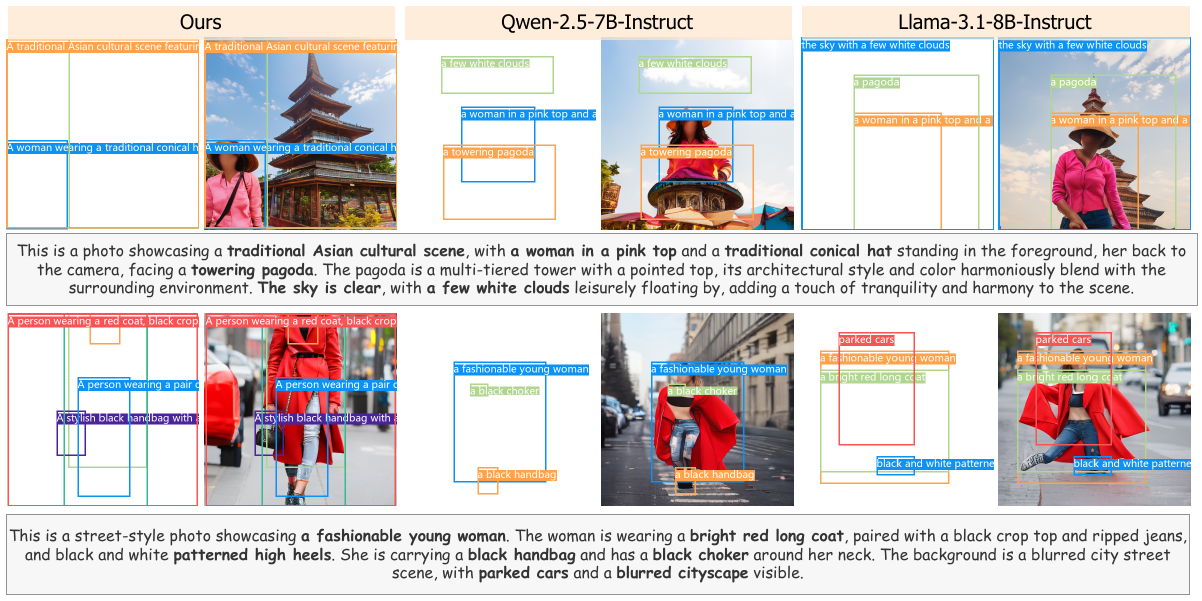}
\caption{Examples for layout planning. Compared with Qwen-2.5-7b-instruct~\cite{qwen2.5-7b-it} and Llama-3.1-8b-instruct~\cite{llama3.1}, \ours{} generates more reasonable layout conditions from complex global captions. We use \ours{} to conduct layout-to-image generation for the layout conditions generated by the three methods to further observe the quality of the layout conditions.}
\label{fig:plan_comp}
\end{figure*}

\section{Experiments}
\begin{figure*}[!t]
\centering
\includegraphics[width=7in]{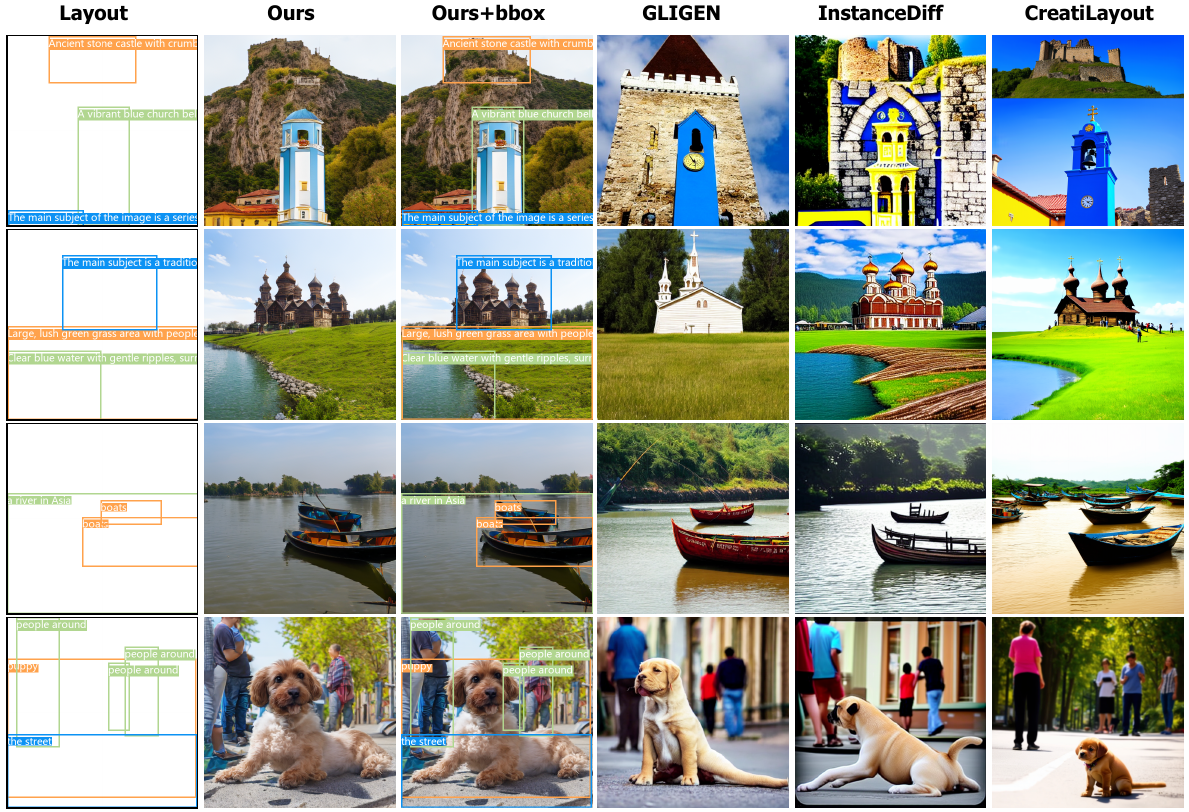}
\caption{Qualitative comparisons of \ours{} with other baselines on layout-to-image generation task, including GLIGEN~\cite{li2023gligen}, InstanceDiff~\cite{wang2024instancediffusion}, and CreatiLayout~\cite{zhang2024creatilayout}. \ours{} generates realistic images that meet the layout conditions, which has obvious advantages compared to previous diffusion-based methods.}
\label{fig:comp}
\end{figure*}

\subsection{Training Datasets}
As shown in Table~\ref{tab:datasets}, we use two existing large-scale layout-to-image datasets in training, including HiCo~\cite{cheng2024hico} derived from GRIT-20M~\cite{peng2023kosmosgrit} and LayoutSAM~\cite{zhang2024creatilayout} derived from SAM~\cite{kirillov2023sam}.
In addition, we also introduce OpenImage V6~\cite{kuznetsova2020open} dataset. We apply MiniCPM~\cite{minicpm-v-2.6} to caption the images and use object class names as local captions, which complements the long local captions in HiCo and LayoutSAM.
Furthermore, considering that these three datasets focus on complex prompts and layouts, we also introduce pure planning data constructed by LayoutGPT~\cite{feng2024layoutgpt}.

\begin{table}[t]
\vspace{-7pt}
\centering
\small
\setlength{\tabcolsep}{2.5pt}
\begin{tabular}{@{}cccccc}
\toprule
\textbf{Dataset}   &  \makecell[c]{{\bf Layout} \\ {\bf Planning}} & \makecell[c]{{\bf Layout-to-} \\ {\bf Image Gen.}}  & \makecell[c]{{\bf Image Lay-} \\ {\bf out Und.}}  & \textbf{\# Samples} \\
\midrule
HiCo~\cite{cheng2024hico}                    & \ding{51}     & \ding{51}             & \ding{51}  & 1,250,466 \\
LayoutSAM~\cite{zhang2024creatilayout} & \ding{51}     & \ding{51}             & \ding{51} & 2,665,509 \\
OpenImage~\cite{kuznetsova2020open} & \ding{51} & \ding{51} &\ding{51}     & 1,743,042  \\
LayoutGPT~\cite{feng2024layoutgpt} &  \ding{51} & \ding{55} &\ding{55}    & 40,481 \\
\bottomrule
\end{tabular}
\caption{%
    \textbf{Statistics of datasets} used for training.
}
\vspace{-12pt}
\label{tab:datasets}
\end{table}

\subsection{Implementation Details}
Our experiment is based on pre-trained Janus-Pro 1.5B~\cite{chen2025janus_pro}.
The training batch sizes of layout-image joint training, image layout understanding, and pure layout planning are 3, 3, and 2 respectively.
We train our \ours{} for 200,000 steps on 8 × 80GB NVIDIA A100 GPUs with AdamW~\cite{loshchilov2017decoupled_adamw} optimizer, and the learning rate is set to 5e-5 with a total batch size of 64.
Images are uniformly processed to 384 × 384 resolution.

\subsection{Layout Planning}

\noindent\textbf{Settings.}
We compare \ours{}'s layout planning ability with existing powerful foundational LLMs: (1) Qwen2.5-7b-instruct~\cite{qwen2.5-7b-it} (2) Llama3.1-8b-instruct~\cite{llama3.1}. We use in-context learning to make LLMs generate layout conditions according to global captions.
We conduct evaluations on \ours{}-1K benchmark, which is derived from LayoutSAM-Eval, used for multitasking evaluation related to layout.
Considering that the layout planning has a high degree of freedom, it is not reasonable to compare the results with ground truth layouts. Following~\cite{zhang2024creatilayout}, we use \ours{} to perform layout-to-image generation for layouts predicted by different methods and indirectly judge the quality of the layout by judging the quality of generated images. We calculate (1) PickScore, which derives from the preference predictor, (2) CLIPScore, which measures the CLIP~\cite{radford2021clip} image-text similarity between the generated images and corresponding text prompts, (3) FID~\cite{heusel2017fid} and (4) Inception Score (IS)~\cite{salimans2016is}.

\noindent\textbf{Results.}
As shown in Figure~\ref{fig:plan_comp}, \ours{} can generate reasonable layouts from complex global captions. Compared with Qwen2.5-7b-instruct and Llama3.1-8b-instruct, the local descriptions of bounding boxes generated by \ours{} are more detailed and specific, which promotes the consistency of the generation results with the global captions. Furthermore, the planned layouts enjoy more reasonable spatial relations, resulting in better image quality.
The quantitative analysis in Table~\ref{tab:plan} also confirms the advantage of \ours{} in layout planning. Compared with the other two LLMs, the image results generated based on \ours{}'s layout predictions have lower FID and are even close to the results generated by using ground-truth layout conditions.

\begin{table*}[]
\centering
\setlength{\tabcolsep}{5pt}
\tabcolsep=0.08cm
\scalebox{1.0}{
\begin{tabular}{@{}clccccccccc@{}}
\toprule
\multirow{2}{*}{Type} & \multirow{2}{*}{Method} & \multicolumn{4}{c}{Region-wise Quality} & \multicolumn{5}{c}{Global-wise Quality} \\ 
\cmidrule(lr){3-6} \cmidrule(l){7-11}
& & Spatial~$\uparrow$   & Color~$\uparrow$   & Texture~$\uparrow$  & Shape~$\uparrow$  & IR~$\uparrow$~\cite{xu2023imagereward}     & Pick~$\uparrow$~\cite{kirstain2023pick}  & CLIP~$\uparrow$  & FID~$\downarrow$ & IS~$\uparrow$  \\ \midrule
& \gray{Real Images}          &   \gray{98.95}	&\gray{98.45}	&\gray{98.90}	 &\gray{98.80}                  & -  &  -  & -  & -  & - \\  \midrule
\multirow{5}{*}{Diffusion} & GLIGEN~\cite{li2023gligen}             & 77.53     & 49.41   & 55.29    & 52.72  & -10.31   & 20.78  & 32.42 & 21.92 & 20.57 \\
& Ranni~\cite{feng2024ranni}               & 41.38     & 24.10    & 25.57    & 23.35  & -28.46  & 20.49  & 31.40  & 27.24 & 19.81 \\
& MIGC~\cite{zhou2024migc}                & 85.66     & 66.97   & 71.24    & 69.06  & -13.72  & 20.71  & 31.36 & 21.19 & 19.65 \\
& InstanceDiff~\cite{wang2024instancediffusion}            & {87.99}     & {69.16}   & {72.78}    & {71.08}  & 9.14   & 21.01  & 31.40  & 19.67 & 20.02 \\
& \hico{}~\cite{cheng2024hico}         & 90.22     & 69.89   & 73.95    & 72.75  & \underline{42.87} & \textbf{22.44}   & \underline{33.11} & 19.30  & \textbf{22.50} \\
& CreatiLayout~\cite{zhang2024creatilayout}         & \textbf{92.67}     & \underline{74.45}   &\underline{77.21}    & \underline{75.93}  & \textbf{69.47}   & \underline{22.02}  & \textbf{34.01} & \underline{19.10}  & \underline{22.04} \\
\midrule
\multirow{1}{*}{AR}  
& \textbf{\ours{}}         & \underline{92.21}     & \textbf{82.69}   & \textbf{86.53}    & \textbf{85.36}  & 39.58   & 21.43  & 31.96 & \textbf{13.91}  & 19.31 \\
\bottomrule
\end{tabular}}
\caption{Layout-to-image quantitative comparison on the LayoutSAM-Eval. Only \ours{} is based on the autoregressive paradigm.}
\label{tab:g2i}
\vspace{-5pt}
\end{table*}

\begin{table}[t]
\vspace{-7pt}
\centering
\begin{tabular}{@{}lccccc}
\toprule
Method & Pick$\uparrow$ & CLIP $\uparrow$ & FID $\downarrow$ & IS$\uparrow$ \\
\midrule
\gray{GT Layout} &\gray{21.44}  &\gray{31.91} &\gray{48.67} &\gray{13.97} \\
\midrule
Qwen2.5-7b-it~\cite{qwen2.5-7b-it}  &20.92  &31.47 &472.38 &13.44 \\
Llama3.1-8b-it~\cite{llama3.1} &21.01  &31.64 &464.07 &13.77 \\
\midrule
\ours{} &\textbf{21.42} & \textbf{31.93}  &\textbf{50.73} & \textbf{15.44}\\
\bottomrule
\end{tabular}
\caption{%
    Layout planning quantitative comparision of \ours{} with Qwen2.5-7b-instruct~\cite{qwen2.5-7b-it} and Llama3.1-8b-instruct~\cite{llama3.1} on \ours{}-1K.
}
\label{tab:plan}
\vspace{-12pt}
\end{table}

\subsection{Layout-to-Image Generation}

\noindent\textbf{Settings.}
Layout-to-image generation was dominated by diffusion-based models. We evaluate \ours{}'s ability of layout-to-image with (1) GLIGEN~\cite{li2023gligen} (2) Ranni~\cite{feng2024ranni} (3) MIGC~\cite{zhou2024migc} (4) InstanceDiff~\cite{wang2024instancediffusion} (5) HiCo~\cite{cheng2024hico} and (6) Creatilayout~\cite{zhang2024creatilayout}.
\ours{} is the only layout-to-image model based on the autoregression paradigm.
We conduct evaluations on LayoutSAM-Eval benchmark following~\cite{zhang2024creatilayout} and report region-wise quality and global-wise quality. The region-wise quality includes spatial, color, texture, and shape scores obtained from MiniCPM~\cite{minicpm-v-2.6}, while the global-wise quality includes traditional indicators like FID, IS, and so on. More details can be found in ~\cite{zhang2024creatilayout}.

\noindent\textbf{Results.}
As shown in Figure~\ref{fig:comp}, \ours{} could generate corresponding images based on given layout conditions, and the generated results show a high degree of consistency with the layout conditions.
In the 1st row, only \ours{} generates a reasonable ``\textit{ancient stone castle}". Although CreatiLayout has a higher image aesthetics and realism, it produces a result of the separation of the upper and lower parts of the image, which is not harmonious.
We observe that the results generated by InstanceDiff lack sufficient realism and tend toward animation style. GLIGEN will ignore certain layout information in some cases, such as the 2nd row, missing ``\textit{clean blue water}".
\ours{} is generally capable of generating realistic images and meeting layout conditions. However, it shows limitations in generating human figures, as evidenced in the last row.
This is consistent with the quantitative results in Table~\ref{tab:g2i}, \ours{} performs well in region-wise quality and significantly exceeds previous diffusion-based methods, thanks to its strong understanding of the layout context. In addition, the FID metric of \ours{} also leads other methods, verifying its excellent image generation quality.



\subsection{Image Layout Understaning}
\noindent\textbf{Settings.}
Image layout understanding has similarity to grounding detection to some extent, both to obtain local captions and corresponding bounding box coordinates, but image layout understanding does not aim at detecting objects in the figure as its ultimate goal.
We compare \ours{}'s image layout understanding ability with specialized grounding detection model (1) \gdino{}~\cite{Liu2023GroundedSAMdino} and LLMs with grounding capability (2) \qwenvl{}~\cite{Qwen-VL} (3) \cogvlmg{}~\cite{wang2023cogvlm}.
We conduct the evaluation on our \ours{}-1K benchmark, reporting AP, AP50, AP75, and AR.
Additional details are provided in the supplemental material.

\noindent\textbf{Results.}
As shown in Table~\ref{tab:mmu}, compared with \qwenvl{} and \cogvlmg{}, \ours{}, specially trained in image layout understanding, achieves excellent layout understanding capabilities with higher AP and AR metrics. It should be noted that \ours{} can even be close to the performance of \gdino{}, which is often a specialized tool for data annotation, showing \ours{}'s promising prospects.

\begin{table}[!t]
\centering
\setlength{\tabcolsep}{3.2pt}
\begin{tabular}{lcccc}
\toprule
Method & AP $\uparrow$ & AP50 $\uparrow$  & AP75 $\uparrow$ & AR $\uparrow$\\
\midrule
\gray{\gdino{}~\cite{Liu2023GroundedSAMdino}} & \gray{50.00} & \gray{61.66} & \gray{54.92} & \gray{69.34} \\
\qwenvl{}~\cite{Qwen-VL} & 5.16 & 9.17 & 5.82 & 20.64 \\
\cogvlmg{}~\cite{wang2023cogvlm} & 9.05 & 13.03 & 9.95 & 29.27\\
\midrule
\ours{} &\textbf{33.48} &\textbf{44.76}  &\textbf{36.71} &\textbf{50.98}\\
\bottomrule
\end{tabular}
\caption{Image layout understanding quantitative comparison of \ours{} with Grounding-DINO~\cite{Liu2023GroundedSAMdino}, \qwenvl{}~\cite{Qwen-VL} and \cogvlmg{}~\cite{wang2023cogvlm} on \ours{}-1K.}
\label{tab:mmu}
\vspace{-10pt}
\end{table}

\subsection{Image Manipulation}
\noindent\textbf{Settings.}
In this part, we focus on comparing the object deletion ability of \ours{}. We choose two classic diffusion-based baselines: (1) Instruct-Pix2Pix~\cite{brooks2023instructpix2pix} (2) SD-Inpainting~\cite{rombach2022stablediffusion}, which is specially trained inpainting model, because it is difficult to manipulate the latent directly in the diffusion model, while \ours{} could conduct image generation and editing in one unified model. We report the following indicators in 200 object removal cases derived from COCO~\cite{lin2014coco} test set, including (1) Success Rate (SR), we calculate it by asking the visual language model MiniCPM~\cite{minicpm-v-2.6} whether there are objects to be deleted in targeted area. (2) LocalCLIP, which measures the deletion ability by evaluating the image-text similarity of the local image areas and the corresponding local descriptions. (3) IS, which measures the quality of the image.

\begin{figure}[!t]
\centering
\includegraphics[width=3.2in]{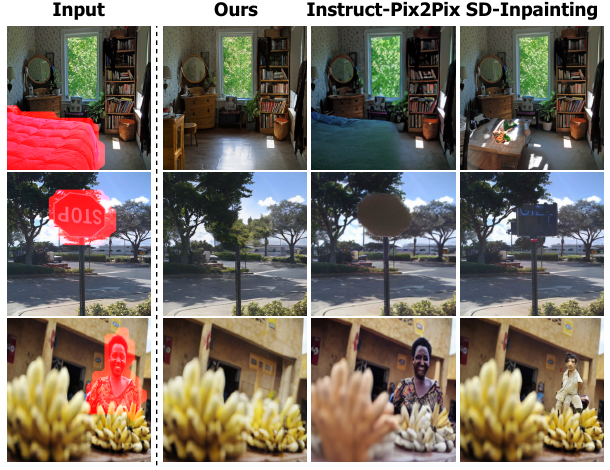}
\caption{Qualitative comparisions of \ours{} with Instruct-Pix2Pix~\cite{brooks2023instructpix2pix} and SD-Inpainting~\cite{rombach2022stablediffusion} on the object removal task. The red areas of images in the 1st column are the objects to be deleted.}
\label{fig:rm}
\vspace{-10pt}
\end{figure}

\noindent\textbf{Results.}
As shown in Figure~\ref{fig:rm}, \ours{} has more obvious advantages when deleting objects, and can more thoroughly delete objects in local areas, making it less likely to leave artifacts or cause image quality to decline. This is consistent with the results in Table~\ref{tab:rm}, and \ours{} achieves the highest deletion success rate.
As illustrated in the first row of Figure~\ref{fig:rm}, when the bed is removed, \ours{} also generates a light reflection on the floor originating from the window.
Instruct-Pix2Pix often causes most of the entire image to change, and it is often impossible to completely delete. SD-Inpainting is prone to leaving artifacts in the area to be deleted, affecting the quality of the image.
Additional image manipulation examples for more editing types are provided in the supplemental material.

\begin{table}[!t]
\centering
\setlength{\tabcolsep}{7.2pt}
\begin{tabular}{lccc}
\toprule
Method & SR $\uparrow$ & LocalCLIP $\downarrow$  & IS $\uparrow$\\
\midrule
Instruct-Pix2Pix~\cite{brooks2023instructpix2pix} & 40.00 & 23.65 & \textbf{9.20}  \\
SD-Inpainting~\cite{rombach2022stablediffusion} & {66.50} & {22.61} & 7.73 \\
\midrule
\ours{} &\underline{77.50} &\underline{21.96}  &\underline{8.45}\\
\small{+Neg. Layout Guidance} &\textbf{86.50} &\textbf{21.90}  &{8.43}\\
\bottomrule
\end{tabular}
\caption{Object removal quantitative comparison of \ours{} with Instruct-Pix2Pix~\cite{brooks2023instructpix2pix} and SD-Inpainting~\cite{rombach2022stablediffusion}.}
\label{tab:rm}
\end{table}

\subsection{Ablation Study}

\begin{table}[!t]
\centering
\small
\setlength{\tabcolsep}{3.2pt}
\begin{tabular}{cccc|cccc}
\toprule
LIG & LP & ILU & Enc & Spatial $\uparrow$ & Color $\uparrow$  & Texture $\uparrow$ & Shape$\uparrow$ \\
\midrule
\ding{51} &\ding{55}  &\ding{55} & N & \underline{87.19} & \textbf{72.25} & \textbf{77.18} & \textbf{76.35}\\
\ding{51} &\ding{51}  &\ding{55} & N & 83.25 & \underline{69.46} & 73.73 & 70.28\\
\ding{51} &\ding{51}  &\ding{51} & N & \textbf{87.52} & 68.14 & \underline{74.55} & \underline{71.26}\\
\ding{51} &\ding{51}  &\ding{51} & S & 81.94 & 59.61 & 67.65 & 66.34\\
\bottomrule
\end{tabular}
\caption{Ablation studies about the task types used in \ours{}'s training and the encoding type of numerical values (Enc). ``LIG'' stands for Layout-to-Image Generation, ``LP'' for Layout Planning, and ``ILU'' for Image Layout Understanding. ``S'' refers to encoding numerical values into special tokens, whereas ``N'' means treating numerical values as normal text.}
\label{tab:abla}
\vspace{-10pt}
\end{table}

In Table~\ref{tab:abla}, we perform a series of ablation experiments on a subset of the training dataset. we compare the different encoding types of numerical values and find that \ours{} can achieve better results by directly considering the values as texts rather than encoding them into special tokens.
In addition, we also ablate the impact of different tasks adopted in the experiment.
The introduction of image layout understanding can improve the effect of the model.
The model trained with all tasks has a similar effect as the model trained with layout-to-image task alone, which verifies the effectiveness of \ours{} using multi-task training.
In addition, in Table~\ref{tab:rm}, we ablate the impact of whether negative layout guidance is used on the object deletion task. By introducing this strategy, we observe that the deletion success rate has been greatly improved.

\section{Discussion}
In this work, we demonstrate a unified model that integrates layout planning and layout-to-image generation into an autoregressive vision-language model, which also has the ability to understand the layout of real images. It performs well in layout planning, layout-to-image, and image layout understanding tasks, showing the great potential of a unified model. In addition, it can be expanded into image manipulation without further training with teacher-forcing content manipulation and negative layout guidance.

While promising, our work still faces several challenges, firstly, our experiments are trained on image resolution of 384×384 due to limited training resources. Experiments at higher resolutions will help further improve the model's effect. In addition, subsequent work can consider more advanced autoregressive image generation paradigms, including mask autoregressive generation~\cite{chang2022maskgit} and next-scale prediction~\cite{VAR} paradigms.

{
    \small
    \bibliographystyle{ieeenat_fullname}
    \bibliography{main}
}

\clearpage
\appendix

\section{Inference Details}
We used classifier-free guidance during the inference process of image generation, and the guidance scale is 5.
We only apply negative layout gudance when performing layout-guided image manipulation including object deletion.
To speed up the inference, we use kv-cache. It takes about 17 seconds to generate 8 images in one batch on a single A100 GPU.

\section{Experimental Details}

\noindent \textbf{In-context Prompt for Baselines in Layout Planning.}
Qwen-2.5-7b-instruct and Llama-3.1-8b-instruct themselves do not naturally support the layout planning task. We leverage LLMs' in-context learning capabilities to generate layouts from global captions following LayoutGPT, and the in-context prompts are as follows:

\begin{lstlisting}[basicstyle=\ttfamily, breaklines=true, linewidth=1.0\columnwidth]
You are an intelligent bounding box generator. I will provide you with a caption for a photo, image, or painting. Your task is to generate the bounding boxes for the objects mentioned in the caption, along with a background prompt describing the scene. The images are of size 512x512. The top-left corner has coordinate [0, 0]. The bottom-right corner has coordinnate [512, 512]. The bounding boxes should not overlap or go beyond the image boundaries. Each bounding box should be in the format of (object name, [top-left x coordinate, top-left y coordinate, box width, box height]) and should not include more than one object. Do not put objects that are already provided in the bounding boxes into the background prompt. Do not include non-existing or excluded objects in the background prompt. Use "A realistic scene" as the background prompt if no background is given in the prompt. If needed, you can make reasonable guesses. Please refer to the example below for the desired format.

input: A realistic image of landscape scene depicting a green car parking on the left of a blue truck, with a red air balloon and a bird in the sky
you need output: [('a green car', [21, 281, 211, 159]), ('a blue truck', [269, 283, 209, 160]), ('a red air balloon', [66, 8, 145, 135]), ('a bird', [296, 42, 143, 100])]

input: A realistic top-down view of a wooden table with two apples on it
you need output: [('a wooden table', [20, 148, 472, 216]), ('an apple', [150, 226, 100, 100]), ('an apple', [280, 226, 100, 100])]

input: An oil painting of a pink dolphin jumping on the left of a steam boat on the sea
you need output: [('a steam boat', [232, 225, 257, 149]), ('a jumping pink dolphin', [21, 249, 189, 123])]

The input for you to process is: {}
\end{lstlisting}

\begin{figure*}[!t]
\centering
\includegraphics[width=7in]{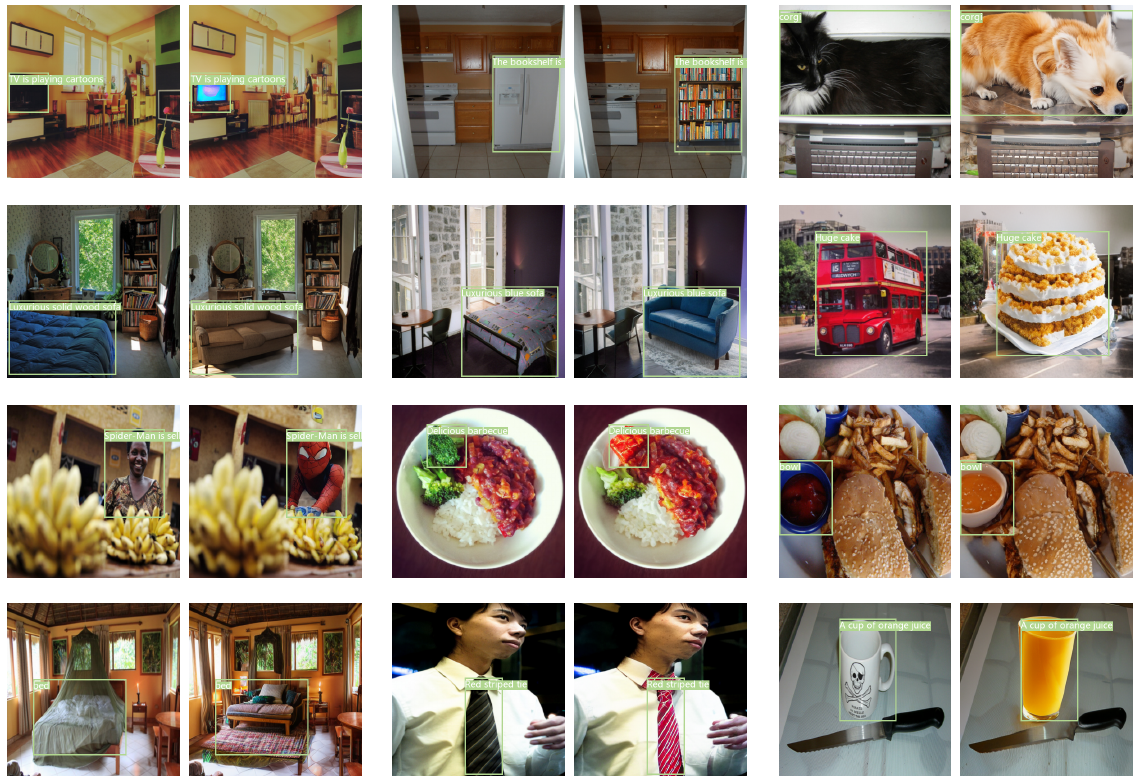}
\caption{More examples for Layout-guided Image Manipulation. The contents to be edited are drawn in the form of bounding boxes on the original images and the edited images for easy comparisons.}
\label{fig:edit}
\vspace{-10pt}
\end{figure*}

\begin{figure}[!t]
\centering
\includegraphics[width=3.5in]{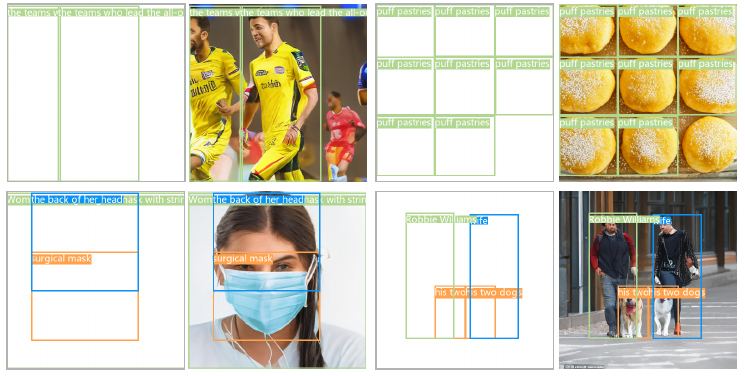}
\caption{Failure cases.}
\label{fig:fail}
\vspace{-15pt}
\end{figure}

\noindent \textbf{Baselines Details on Image Layout Understanding.}
For Grounding-DINO, we perform grounding detection by giving the image and the global caption of the image. For two other LLM-based baselines, i.e. CogVLM-grounding and Qwen-VL-Chat, we take prompts suitable for the corresponding models to help the model to output accurate detection results.
Specifically, for CogVLM-grounding, we give the image and ask ``\textit{Can you provide a description of the image and include the coordinates [[x0,y0,x1,y1]] for each mentioned object?}'' following its formal demo. For Qwen-VL-Chat, we apply two rounds of questions. First, we give the image and ask the model ``\textit{What objects are in the image?}''. Then, after the model answers this question, we ask the model ``\textit{Box out the positions of these objects in the figure}''. We find that for Qwen-VL-Chat, the effect of such two-round Q\&A will be much better than a single question.

\noindent \textbf{Evaluation of Image Layout Understanding.}
Similar to HiCo, we calculate the maximum IoU between the boxes predicted by different models and the ground truth boxes. If the maximum IoU is higher than the threshod 0.5, we caculate the clip text similarity between corresponding local descriptions. If the CLIP score is higher than 0.2, we mark it as a correct prediction. We use AR, AP, AP50 and AP75 to evaluate the performance of image layout understanding.

\vspace{-5pt}
\section{Additional Results}
\begin{figure*}[!t]
\centering
\includegraphics[width=6.2in]{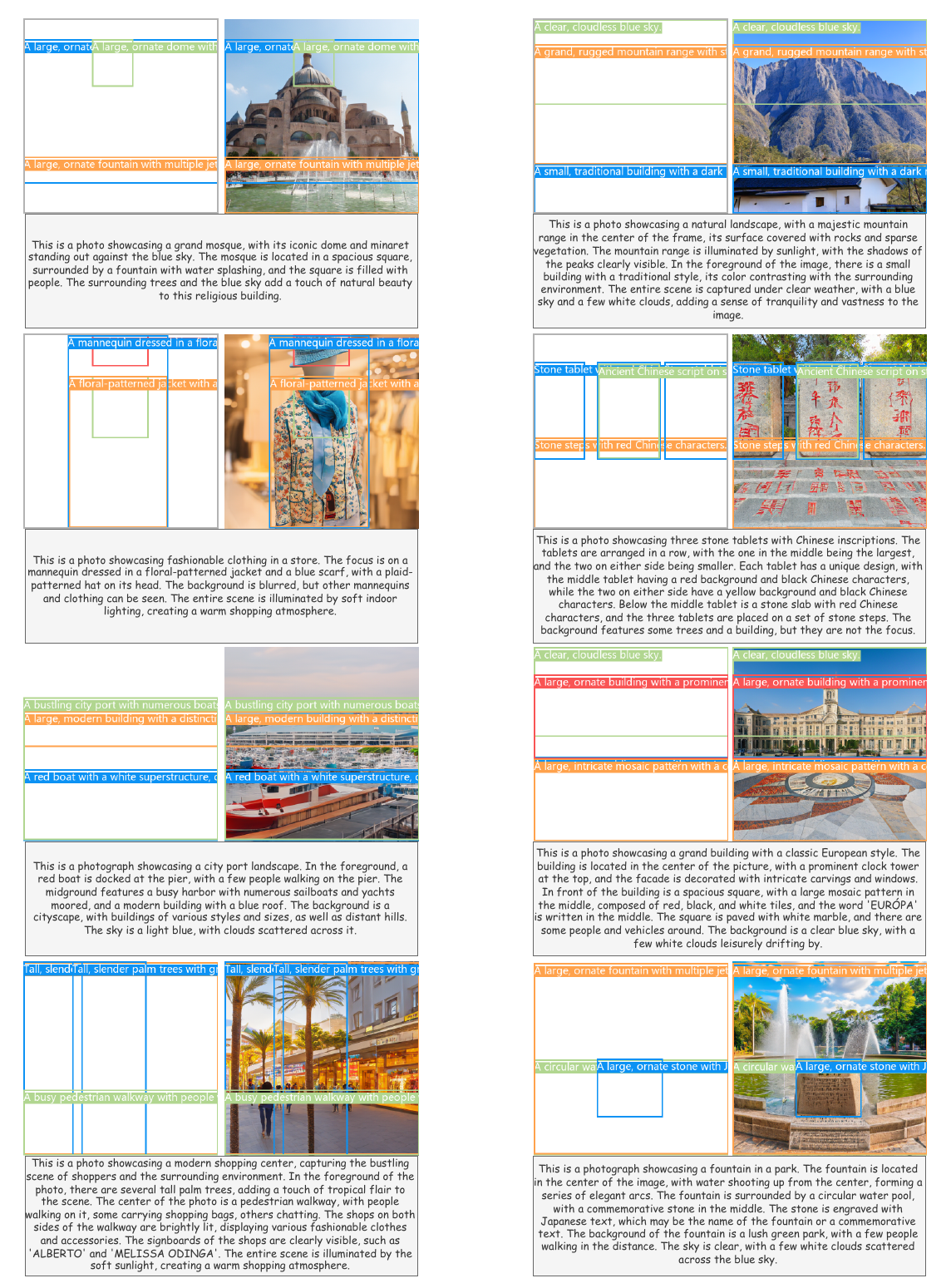}
\caption{More examples for Layout-Image Joint Generation. Global captions for layout-image joint generation are attached below the images.}
\label{fig:m_2stage}
\vspace{-10pt}
\end{figure*}

\begin{figure*}[!t]
\centering
\includegraphics[width=7in]{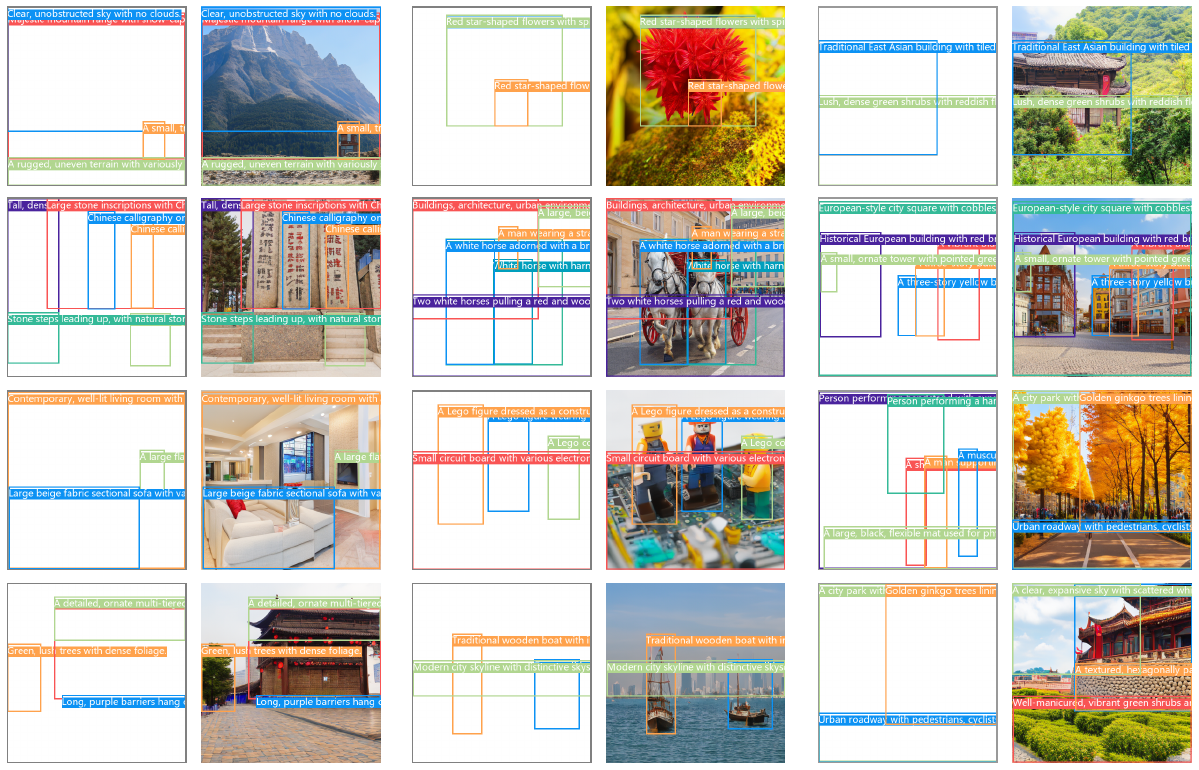}
\caption{More examples for Layout-to-Image Generation.}
\label{fig:m_g2i}
\end{figure*}

\begin{figure*}[!t]
\centering
\includegraphics[width=7in]{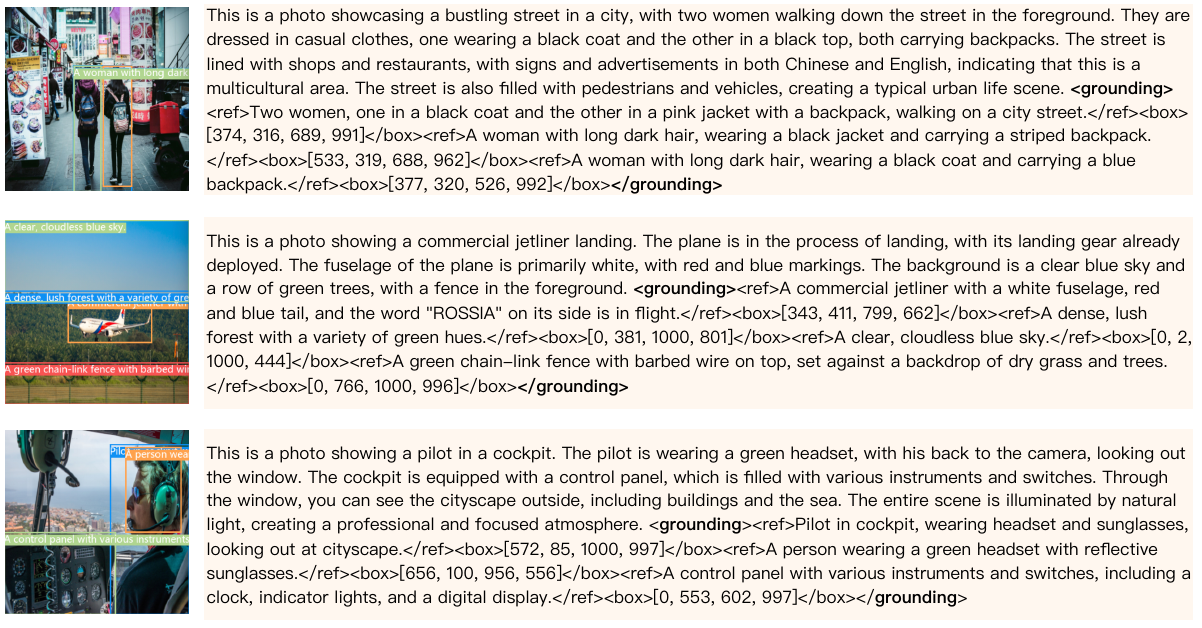}
\caption{More examples for Image Layout Understanding.}
\label{fig:m_mmu}
\vspace{-10pt}
\end{figure*}

We show more examples of layout-guided image manipulation in Figure~\ref{fig:edit}, more examples of layout-image joint generation in Figure~\ref{fig:m_2stage}, more examples of layout-to-image in Figure~\ref{fig:m_g2i}, and more results of image layout understanding in Figure~\ref{fig:m_mmu}.
These rich examples show PlanGen's excellent performance on multiple related tasks.

\noindent \textbf{Failure cases.}
We also show several failure cases on the task of layout-image joint generation in Figure~\ref{fig:fail}.
We observed that PlanGen may experience distortion when generating human bodies, as shown in Figure~\ref{fig:fail}, which is a common challenge faced by some previous autoregressive image generation models. When multiple identical objects are generated, additional ones may appear, which is caused by incomplete object annotations in the training data. In the 2nd row, the the geometry of generated mask straps is slightly inappropriate. When generating dogs in smaller size, the quality declines also occur. More efficient image modeling or wider training should alleviate these problems.

\end{document}